\newcommand{\CLASSINPUTtoptextmargin}{20.6mm}
\newcommand{\CLASSINPUTbottomtextmargin}{25.5mm}
\documentclass[10pt, conference, compsocconf]{IEEEtran}
\usepackage[utf8]{inputenc}
\usepackage[T1]{fontenc}
\usepackage{cite}
\usepackage{amsmath,amssymb,amsfonts}
\usepackage{graphicx}
\usepackage{textcomp}
\usepackage{xcolor}

\usepackage[compact]{titlesec}
\titlespacing{\section}{0pt}{1ex}{1ex}
\titlespacing{\subsection}{0pt}{1ex}{0ex}
\titlespacing{\subsubsection}{0pt}{0.5ex}{0ex}
\usepackage{comment}
\usepackage{tabularx}
\usepackage{booktabs}
\usepackage{multirow}
\usepackage{paralist}
\usepackage{algorithm}
\usepackage{algpseudocode}
\usepackage{upgreek}
\DeclareMathOperator*{\argmax}{arg\,max}

\usepackage[singlespacing]{setspace} 
\usepackage{etoolbox}
\usepackage[font=small]{caption}
\makeatletter
\newcommand{\StatexIndent}[1][3]{%
  \setlength\@tempdima{\algorithmicindent}%
  \Statex\hskip\dimexpr#1\@tempdima\relax}
\makeatother
\makeatletter
\patchcmd{\@makecaption}
  {\scshape}
  {}
  {}
  {}
\makeatletter
\patchcmd{\@makecaption}
  {\\}
  {.\ }
  {}
  {}
\makeatother
\def\tablename{Table}
\def\BibTeX{{\rm B\kern-.05em{\sc i\kern-.025em b}\kern-.08em
    T\kern-.1667em\lower.7ex\hbox{E}\kern-.125emX}}

\begin{document}

\title{Privacy-preserving Speech Emotion Recognition through Semi-Supervised Federated Learning}

\author{
    \IEEEauthorblockN{
        Vasileios Tsouvalas\IEEEauthorrefmark{1},
        Tanir Ozcelebi\IEEEauthorrefmark{2} and
        Nirvana Meratnia\IEEEauthorrefmark{3}
    }
    \IEEEauthorblockA{
        Department of Mathematics and Computer Science,
        Eindhoven University of Technology\\
        Email:  \IEEEauthorrefmark{1}v.tsouvalas@tue.nl,
                \IEEEauthorrefmark{2}t.ozcelebi@tue.nl,
                \IEEEauthorrefmark{3}n.meratnia@tue.nl
    }
}
\maketitle

\begin{abstract}
Speech Emotion Recognition (SER) refers to the recognition of human emotions from natural speech. If done accurately, it can offer a number of benefits in building human-centered context-aware intelligent systems. Existing SER approaches are largely centralized, without considering users' privacy. Federated Learning (FL) is a distributed machine learning paradigm dealing with decentralization of privacy-sensitive personal data. In this paper, we present a privacy-preserving and data-efficient SER approach by utilizing the concept of FL. To the best of our knowledge, this is the first federated SER approach, which utilizes self-training learning in conjunction with federated learning to exploit both labeled and unlabeled on-device data. Our experimental evaluations on the IEMOCAP dataset shows that our federated approach can learn generalizable SER models even under low availability of data labels and highly non-i.i.d. distributions. We show that our approach with as few as 10\% labeled data, on average, can improve the recognition rate by 8.67\% compared to the fully-supervised federated counterparts.
\end{abstract}

\begin{IEEEkeywords}
deep learning, emotion classification, federated learning, semi-supervised learning, speech emotion recognition
\end{IEEEkeywords}

\section{Introduction}
Speech Emotion Recognition (SER) has attracted growing attention due to its beneficial role in building human-centered context-aware intelligent systems in many fields, such as customer support call review and analysis~\cite{app:SER_call}, mental health surveillance~\cite{app:SER_bipolar1}, multimedia retrieval~\cite{app:SER_multimedia} and smart vehicles~\cite{app:SER_vehicles}. The main task of SER is to automatically recognize the human emotional states by analyzing utterances. Since, linguists' emotion sets often exceed 300 states, researchers have agreed to use the ‘palette theory’ to compose a number of core emotions (i.e., Anger, Disgust, Fear, Joy, Sadness, and Surprise) from which any emotional state is originated~\cite{SER:hourgrass}.\par

For many years, SER approaches has focused on recognizing these core emotions by extracting frame-based features from utterances followed by a classification or regression algorithm, such as Support Vector Machines (SVMs)~\cite{SER:old_survey}. With the advent of deep learning, several deep learning models have been proposed to perform SER by learning from raw audio data directly~\cite{SER:first_DNN}. In particular, the recent success of neural network-based attention mechanisms (AMs) across various machine learning tasks, such as image classification~\cite{IC:Attention}, automatic speech recognition~\cite{ASR:Attention} and text translation~\cite{Transformer}, has attracted a growing interest in SER~\cite{SER_Attention:Survey} community. Firstly introduced in the Natural Language Processing (NLP) field~\cite{Attention}, AMs “\textit{help}” the model learn where to “\textit{look for}” information that is meaningful for performing a particular classification task. Extending this concept to SER, the AM can help determine the emotional state of the speaker, by focusing on specific parts of a spoken utterance that contain emotional information, while disregarding noisy or irrelevant data.\par

A common limitation of existing SER approaches is their centralized approach, with the implicit assumption that audio data originated from distributed devices can be aggregated in a centralized repository before performing further processing. However, the rapidly increasing size of available data, in combination with the high communication costs, and possible bandwidth limitations, render the accumulation of data in a cloud server unfeasible~\cite{FL_Challenges}. Additionally, such centralized data processing schemes do not consider the privacy concerns and regulations like the General Data Protection Regulation. These limitations and the growing computational and storage capabilities of distributed devices make it appealing to perform SER directly on the device that collects the data by  utilizing local computational resources and local learning models.\par
Federated Learning (FL) is particularly suitable for this purpose thanks to its unique characteristic of collaboratively training machine learning models without sharing local data and compromising users' privacy~\cite{FL}. The most popular paradigm in FL is the Federated Averaging (FedAvg) algorithm~\cite{FedAvg}, in which minimal updates to local models are performed entirely on-device and only a few model parameter updates are communicated to the central server to aggregate all updates to produce a unified global model. This strategy has recently been applied to a wide range of acoustic tasks~\cite{KeywordSpottingIID, KeywordSpottingNonIID, KeywordSuggestion}. Motivated by their success, in this paper, we investigate the feasibility of SER using FL aiming at preserving users' privacy. Applying traditional FL approaches to SER is undesirable since these FL approach assume that on-device labeled data is sufficiently available, or data can be easily labeled through user interaction or labeling functions, as in keyword prediction. This is clearly an unrealistic assumption in SER. In practice, the ever-growing on-device speech data is largely unlabeled due to the prohibitive cost of annotation and little to no incentives (or expertise) for users to label their data~\cite{SSFL}. \par

To utilize on-device unlabeled data, semi-supervised learning (SSL) under federated settings can be explored. Our hypothesis is that this will significantly boost SER models' performance~\cite{FedSTAR}. In SSL, size of the labeled data is generally much smaller than the unlabeled data~\cite{SSL}. While there is a wide range of SSL methods, we focus on self-training or pseudo-labeling approach~\cite{PSL}. In particular, self-training uses the prediction on unlabeled data to supervise the model's training in combination with a small percentage of labeled data~\cite{PSL}. This simple, yet, effective approach has been shown to achieve great results in centralized regimes~\cite{DeepPSL, DistillingPSL}. In contrast with various data augmentation techniques proposed for SER to increase the training data~\cite{SER:Augmentation}, our FL approach utilizes largely available on-device unlabeled data. This way, we boost the data availability, without losing the emotional content through unfitting deformation on existing data.\par

In this regard, we propose a privacy-preserving and data-efficient SER approach through federated self-training, which unifies semi-supervision with federated learning. By doing so, it addresses significant challenges regarding scarcity of data labels and privacy regulations faced by SER. As distributed devices often operate under energy and computation constraints, we utilize a spectro-temporal-channel attention mechanism to effectively capture relationships across the different segments of utterances without increasing the model's complexity, thus accelerating the training process. To the best of our knowledge, this work is the first federated SER approach that learns models by utilizing not only labeled but also unlabeled samples on user devices, while not being dependent on any data (labeled or unlabeled) on the server side. To this end, our main contributions are as follows:
\begin{itemize}
    \item We present a privacy-preserving and data-efficient SER approach based on semi-supervised federated learning, which exploits both labeled and unlabeled on-device data.
    \item We introduce an attention mechanism to improve the representation power of SER model without increasing its complexity.
    \item We demonstrate, through extensive evaluations on the IEMOCAP dataset~\cite{ds:IEMOCAP}, that our approach effectively learns generalizable SER models under federated settings, even in case of low availability of labeled data.
    \item We show that our approach with as few as 10\% labeled data, on average, can improve the recognition rate by 8.67\% compared to a fully-supervised federated regime.
\end{itemize}

\section{Related Work}

Authors of~\cite{RW0} first investigated the use of deep learning in SER by utilizing a Bi-directional Long Short-Term Memory (BLSTM) architecture to capture time-related relationships across data. This was followed by several papers that showed the contribution of attention models in the SER field. 
Authors of~\cite{RW9} used a BLSTM architecture in conjunction with a local attention scheme, which replaced mean pooling with a weighted time-pooling algorithm to compute a weighted sum of the attention outputs. In~\cite{RW10} a similar AM on top of a CNN model, showing that convolutions solve SER tasks with similar performance. 
Along this path,~\cite{RW1} utilized two types of convolution filters for extracting time-specific and spectral-specific features from spectrograms, after which a CNN architecture for modelling high-level representation was used. Subsequently, to further improve SER performance, ~\cite{RW1} proposed an attention-based poling method, combining two attention maps — a class-specific top-down and a class-agnostic bottom-up attention — on top of the spectral-time feature extraction, to replace Global Average Pooling (GAP). Recently,~\cite{RW7} proposed an attention mechanism, which considers the temporal, spectral, and channel (STC) dimensions simultaneously to tackle the limited ability of CNNs to capture relative importance of features across all 3 axes. With respect to model inputs, all previous works relied on segmenting the utterance into smaller uniformly-shaped chunks, which are then fed to the model. Alternatively,~\cite{RW3} designed a fully convolutional network (FCN) architecture, which can preserve the information from variable length utterances as a whole without the need for segmentation. To further distinguish between important and non-speech parts of the input, an attention mechanism, similar to~\cite{RW10}, was used before the classifier. Apart from the attention-based approaches, transfer learning~\cite{SOTA5} and various augmentation techniques~\cite{SER:Augmentation} have been developed to deal with the limited amount of available natural speech data. None of the existing SER models considers the users' privacy challenge, which can significantly affect their applicability in real-life applications. 
The requirement to preserve users' privacy often results in a limited amount of natural speech data, which augmentations can only increase to a certain extent before deforming the emotional information present in those data.

\textbf{Semi-Supervised Federated Learning:} To realistically move away from a centralized approach to a federated approach, the unrealistic assumption of existing FL approaches that labeled data are largely available on devices needs to be eliminated~\cite{SSFL}. Existing semi-supervised federated learning (SSFL) approaches have only recently started to be examined in the vision domain to exploit unlabeled data~\cite{FedMatch, FedSemi}. FedMatch~\cite{FedMatch} uses an inter-device consistency loss to enforce consistency between the pseudo-labeling predictions made across multiple devices. In~\cite{FedSemi}, FedSemi adapts a mean teacher approach to harvest the unlabeled data during the training process. Nevertheless, none of the discussed approaches focuses on learning audio models by utilizing devices' unlabeled audio samples. Furthermore, these approaches also introduce additional communication overhead to utilize the available on-device unlabeled data.

\section{Methodology\label{sec:methodology}}

\begin{figure*}[t]
    \centerline{\includegraphics[width=1.59\columnwidth]{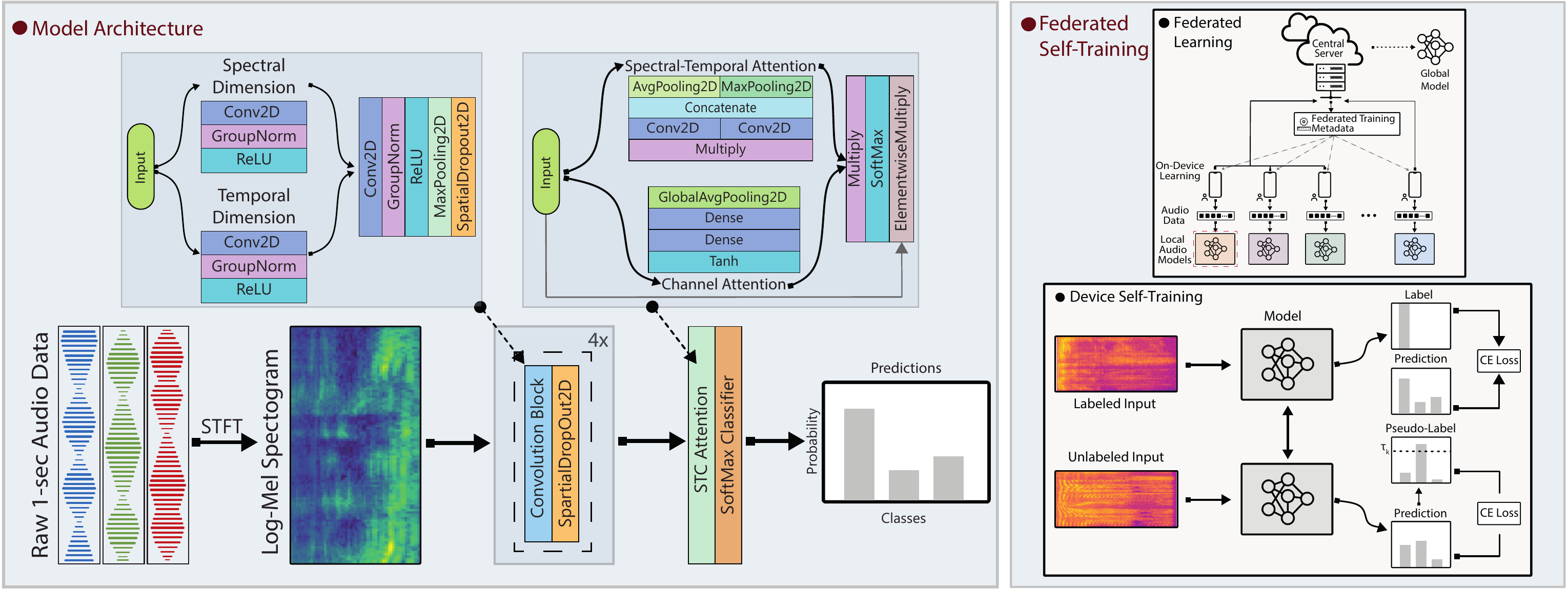}}
    \caption{Proposed attention-based CNN architecture and Federated Self-Training\label{fig:model}}
    \vspace*{-5mm}
\end{figure*}

\textbf{Problem Formulation}:
Formally, when performing SER in a semi-supervised FL setting, each of the $K$ devices holds a labeled set, $\mathcal{D}_{L}^{k} = \{ \left ( x_{l_i},y_{i} \right ) \}_{i=1}^{N_{l,k}}$, where $N_{l,k}$ is the number of labeled data samples, $x_{l_i}$ is an input instance, $y_{i} ~\epsilon \left \{ 1, \cdots, \mathcal{C} \right \}$ is the corresponding label, and $\mathcal{C}$ is the number of label categories for the $\mathcal{C}$-way multi-class classification problem. Besides, each device holds a set of unlabeled samples denoted by $\mathcal{D}_{U}^{k} = \{x_{u_i}\}_{i=1}^{N_{u,k}}$, where, $N_{u,k}$ is the number of unlabeled data samples. Here, $N_{k} = N_{l,k} + N_{u,k}$ is the total number of data samples stored on the $\mathit{k}$-th device and $N_{l,k} \ll N_{u,k}$. Let $p_\theta\left(y \mid x \right)$ be a neural network that is parameterized by weights $\theta$ that predicts softmax outputs $\widehat{y}$ for a given input $x$. We aim to learn a global unified model $G$ without devices sharing any of their local data, $\mathcal{D}_L^k$ and $\mathcal{D}_U^k$. Our objective is to simultaneously minimize both supervised and unsupervised learning losses during each device's local training step on the $\mathit{r}$-th round of the FL algorithm. Specifically, the minimization function is:
\begin{equation} \label{eqn:SSFL}
    \resizebox{0.90\hsize}{!}{%
    $\min_{\theta} {\mathcal{L}}_{\theta} = \sum_{k=1}^{K} \gamma_{k} {\mathcal{L}}_k(\theta) \textrm{ where } \mathcal{L}_{k}(\theta) =\mathcal{L}_{s_{\theta}}(\mathcal{D}_{L}^{k}) + \beta \mathcal{L}_{u_{\theta}}(\mathcal{D}_{U}^{k})%
    $}
\end{equation}

\noindent, where $\mathcal{L}_{s_{\theta}}(\mathcal{D}_{L}^{k})$ is the loss term from supervised learning on the labeled data held by the $k$-th device, and $\mathcal{L}_{u_{\theta}}(\mathcal{D}_{U}^{k})$ represents the loss term from unsupervised learning on the unlabeled data of the same device. We add the parameter $\beta$ to control the effect of unlabeled data on the training procedure, while $\gamma_{k}$ is the relative impact of the $k$-th device on the construction of the global model $G$. For the FedAvg algorithm, the parameter $\gamma_{k}$ is equal to the ratio of device's local data $N_k$ over all training samples $\left (\gamma_{k} = \frac{N_k}{N}\right)$.

\textbf{Model Architecture}:
We use log-Mel spectrograms as the model's input, which we compute by applying a short-time Fourier transform on the two-second audio segment with a window size of $25$ \textit{ms} and a hop size equal to $10$ \textit{ms} to extract $64$ Mel-spaced frequency bins for each window. Inspired by~\cite{Model}, to make a prediction on an audio sample, we average over all the predictions of non-overlapping segments of this sample.\par

Our model architecture consists of four blocks. In each block, we perform two separate convolutions (i.e.,  one on the temporal and another on the frequency dimension), whose outputs we concatenate afterward to perform a joint $1\times1$ convolution. By doing this, the model can capture fine-grained features from each dimension and learn high-level features from their shared output. To accelerate the training process, while avoiding overfitting, we employ group normalization and spatial dropout after each convolution layer. Furthermore, we apply L$2$ regularization with a rate of $0.0001$ in each convolution layer and utilize max-pooling to reduce the time-frequency dimensions by a factor of two between blocks, effectively reducing the number of models parameters. In addition to max-pooling, a spatial dropout rate of $0.1$ was used to avoid further over-fitting. We apply ReLU as a non-linear activation function and use Adam optimizer with the learning rate of $0.001$ to optimize categorical cross-entropy loss.\par

\textbf{\textit{Attention Mechanism}}:
As emotional cues can be sparsely scattered across natural speech utterances, we utilize a spectro-temporal-channel attention mechanism, inspired by~\cite{RW7}, in conjunction with our CNN blocks to improve the representational power of our model without increasing its complexity. The STC attention mechanism consists of two major components, a spectro-temporal (ST) attention to capture prosody (e.g., rhythm, pitch, and intonations) and spectral (e.g., formants and harmonics) patterns, and a channel attention to discover interactions across CNN channels. \par
The channel attention is obtained by first aggregating the features from each channel though GAP, and then passing them through a two-layer perceptron to construct the channel attention map, where we adjust its scale using the \textit{tanh} function. Concurrently, the spectro-temporal (ST) attention is computed by concatenating two feature maps extracted from average pooling and max pooling along the channel dimension, in which we perform two separate convolutions across both spectral and temporal dimensions. To extract the STC attention weights, the three attention maps are multiplied and scaled using a SoftMax function. We produce the attention-weighted features by element-wise multiplying the extracted STC attention weights with the originally-extracted CNN features, which we then feed to the classifier. Figure~\ref{fig:model} illustrates an overview of our proposed architecture.\par

\textbf{Federated Self-Training for SER}:
Inspired by~\cite{FedSTAR}, we present a federated self-training algorithm, which utilizes unlabeled audio data residing on the devices using a device-specific confidence threshold for predictions. To learn from the labeled datasets $\mathcal{D}_{L}^k$ across all participating devices, we use cross-entropy loss $\mathcal{L}_{s_{\theta}}(\mathcal{D}_{L}^{k}) = \mathcal{L}_{CE}\left ( y,p_{\theta^k}\left(y\mid x_l \right) \right)$. Next, to learn from unlabeled data, we generate pseudo-labels $\widehat{y}$ for unlabeled data $x_{u}$ of device $\mathit{k}$ as:

\vspace*{-3mm}
\begin{equation}\label{eqn:PSL}
        \widehat{y} = \Upphi \left ( z,T \right ) = \argmax\limits_{i \in \left \{ 1, \dots, \mathcal{C} \right \}} \left (\frac{e^{z_i/T}}{\sum_{j=1}^{\mathcal{C}} e^{z_j/T}} \right )
\end{equation}

\noindent, where $z_{i}$ is the logits produced for the input sample $x_{u_i}$ by the $k$-th device model $p_{\theta^k}$ before the softmax layer. In essence, $\Upphi$ produces categorical labels for the given “\textit{soften}” softmax values, in which temperature scaling is applied with a constant scalar temperature $T$. With the help of a confidence threshold, $\tau$, we retain only high-confidence predictions, for which we then perform standard cross-entropy minimization while using $\widehat{y}$ as targets, $\mathcal{L}_{u_{\theta}}(\mathcal{D}_{U}^{k}) = \mathcal{L}_{CE}\left ( \widehat{y},p_{\theta^k}\left(x_u\right) \right )$. \par

In contrast to~\cite{FedSTAR}, to ensure the proper utilization of unlabeled samples in SER, we present a device-specific threshold $\tau$, which follows a cosine scheduler that considers both the total number of rounds being completed and the device's specific participation in the FL procedure. Formally, the device-specific confidence threshold is defined as:

\vspace*{-3mm}
\begin{equation}\label{eqn:tau}
    \resizebox{0.90\hsize}{!}{%
        $\begin{aligned}
            \tau&= ConfidenceSceduler\left (R,C,C_s\right )\\
                 & = \frac{1}{2} \left (\tau_{max}-\tau_{min}\right )\cdot \left (1+cos\left (\frac{1}{R}\cdot
                    \left (C-\delta\cdot
                    \left (C-C_s\right )\right )\cdot
                    \pi\right )\right )
        \end{aligned}$%
    }
\end{equation}

\noindent, where $R$ is the total number of rounds the federated algorithm will be performed, $C$ and $C_s$ are the current total number of completed rounds of the federated procedure globally, and the current number of completed rounds by the specific device, respectively. Additionally, we control the effect of devices' participation in $\tau$ with $\delta$, which we set to 0.5. Thus, we ensure that we utilize solely high-confidence predictions when learning from unlabeled dataset $D_{U}^{k}$, even under low participation rate of devices. Further details and an overview of our approach for the federated self-training procedure can be found in Algorithm~\ref{alg:PSL}.

\begin{algorithm}[t]
    \caption{Federated Self-training for SER. $\eta$ is the learning rate, $l$ and $u$ are equally sized on-device labeled and unlabeled batches, respectively.\label{alg:PSL}}
    \footnotesize
    \begin{algorithmic}[1]
        \State Server initialization of model $G$ with model weights $\theta_{0}^{G}$
        \For{ $i=1,\dots,R$ }
            \State Randomly select $K$ devices to participate in round $i$
            \For{ each device $k \in K$ \textbf{ in parallel}}
                \State $\theta_{i}^{k} \gets \theta_{i}^{G}$
                \State $\theta_{i+1}^{k} \gets$ DeviceUpdate($\theta_{i}^{k}$) 
            \EndFor
            \State $\theta_{i+1}^{G} \gets \sum\nolimits_{k=1}^{K} \frac{N_k}{N} \theta_{i+1}^k$
        \EndFor
        \Procedure{DeviceUpdate}{$\theta$}
            \For{epoch $e=1,2,\dots,E$}
                \For{ batch $l \in \mathcal{D}_{L}$ and $u \in \mathcal{D}_{U}$}
                    \State $\widehat{y} \gets \Upphi \left( p_{\theta}(x_u),T \right)$
                    \State {$\theta \gets \theta - \eta \nabla_{\theta} \left(  \mathcal{L}_{CE}(y,p_{\theta}(y \mid x_l)\right)$}
                    \StatexIndent[8] $+\beta \cdot \mathcal{L}_{CE}\left(\widehat{y},p_{\theta}(x_u))\right)$
                \EndFor
            \EndFor
        \EndProcedure
    \end{algorithmic}
\end{algorithm}

\section{Evaluation\label{sec:evaluation}}

\textbf{Datasets}:
We conducted experiments with the IEMOCAP database~\cite{ds:IEMOCAP}, a widely used dataset across literature~\cite{SER_Attention:Survey}. It contains around 12 hours of audio data, sampled at 16 kHz and is composed of five dyadic sessions where actors perform improvisations or scripted scenarios, specifically to represent the emotional expressions. Since the improvised corpus is closer to natural speech and can elicit more intense emotions, we use only the improvised raw audio samples from the dataset. Additionally, as most papers on SER have targeted the improvised corpus, with a focus on the detection of four core emotion — Neutral, Happy, Sad and Angry~\cite{RW3,RW9,SOTA5}, we use these 4 emotions to be able to compare our results.\par

\textbf{Simulation Environment and Setup}: 
To simulate a federated environment, we use the Flower framework~\cite{Flower} and utilize FedAvg~\cite{FedAvg} as an optimization algorithm to construct the global model from devices' local updates. We select a number of parameters to control the federated settings of our experiments. These parameters are: \begin{inparaenum}[1)]
\item $K$ — number of devices, 
\item $R$ — number of rounds, 
\item $q$ — devices' participation percentage in each round, 
\item $E$ — number of local training steps per round, 
\item $\sigma$ — variance of data distribution across devices,
\item $L$ — dataset's percentage to be used as labeled samples,
\item $\beta$ — influence of unlabeled data over training process, 
\item $T$ — temperature scaling parameter, and
\item $\tau$ — predictions' confidence threshold.
\end{inparaenum} 
Across all semi-supervised experiments, we utilized the federated self-training algorithm, as presented in Section~\ref{sec:methodology}, where we fixed $T=2$ and set $\tau$ to initialize from $0.5$ and gradually increase to a maximum of $0.9$ during training, following Equation~\ref{eqn:tau}.\par 
Data partitioning across devices and into labeled and unlabeled subsets during our semi-supervised experiments plays a key role in performing a realistic evaluation~\cite{SSLRealisticEvaluation}. In our experiments where the creation of a labeled subset from the original is required ($L<$100\%), we keep the dataset's initial class distribution ratio to avoid tempering with dataset characteristics. Likewise, in our semi-supervised federated approach, the unlabeled subset consists of the dataset's remaining samples after extracting the labeled samples.\par

\begin{table}[t]
    \caption{Key Characteristics of Federated Experiments\label{tab:experiment_details}}
    \begin{center}
        \resizebox{\columnwidth}{!}{%
            \begin{tabular}{lccc}
                \toprule
                \multicolumn{1}{c}{\multirow{2}{*}{\textbf{Notation}}} & 
                \multicolumn{2}{c}{\textbf{Train/Test Sets}} &
                \multirow{2}{*}{\textbf{\begin{tabular}[c]{@{}c@{}}Device Data Division\\(Labeled \& Unlabeled)\end{tabular}}} 
                \\ \cmidrule[0.05em](rl){2-3}
                & \begin{tabular}[c]{@{}c@{}}Speakers\\Overlap\end{tabular} &  \begin{tabular}[c]{@{}c@{}}Split\\Strategy\end{tabular} &
                \\ \midrule[0.05em]
                Speaker Indep. (random)& & \textit{LOSO} & Random \\
                Speaker Indep. (non-i.i.d.)   & & \textit{LOSO} & Per Speaker \\
                Speaker Overlap (random)& \textbf{\checkmark} & 5-fold \textit{CV} & Random \\ 
                Speaker Overlap (non-i.i.d.) & \textbf{\checkmark} & 5-fold \textit{CV} & Per Speaker \\
                \bottomrule
            \end{tabular}%
        }
    \end{center}
\end{table}

\textbf{Experiments}:
To evaluate performance of our SER model in a federated setting, we performed four distinct experiments, each of which having a different splitting of on-device data. In particular, based on the placement (or ownership) of the distributed devices, the data partitioning among devices can be either random or in a per-speaker basis. The first corresponds to devices situated in common areas, while the latter to a scenario where devices belong to specific users (e.g., smartphones and computers). Furthermore, SER tasks can also be categorized based on the overlap of speakers between the training and test sets. The case of no overlap of speaker, namely speaker independent SER, has proven especially challenging, since we evaluate the ability of the models to generalize to speakers with different characteristics~\cite{SER_Attention:Survey}. In addition, an overlapping scenario can frequently occur in workplaces (e.g., offices and laboratories) where the number of speakers is fixed. Combining these aforementioned criteria (i.e., placement of distributed devices and speaker overlap between train and test set) results in four federated experiments.\par
To compare our work with recent state-of-the-art approaches, we followed an evaluation strategy similar to~\cite{RW3, RW8, RW9}. In particular, for speaker independent experiments, we utilized leave-one-session-out (\textit{LOSO}) cross-validation strategy. In speaker overlapping experiments, we randomly divide our data into training and test sets using an 80:20 ratio with a 5-fold Cross-Validation (5-fold \textit{CV}) strategy, resulting in 5 distinct train-test sets. It is important to note that since the number of utterances is large enough, each subset will retain the same class distribution as the original data set. We further manage any randomness during the data partitioning and training procedures by passing a specific seed, while we perform five distinct trials (or runs, i.e., training a model from scratch) in each particular splitting, and the average Unweighted Accuracy (\textit{UA}) over all five runs is reported across our results. The key characteristics of our federated experiments are presented in Table~\ref{tab:experiment_details}.\par

\begin{table}[t]
    \caption{UA\% Comparison of Attention-based Models\label{tab:baseline}}
    \begin{center}
        \resizebox{\columnwidth}{!}{%
            \begin{tabular}{lcc}
                \toprule
                \multicolumn{1}{c}{\multirow{2}{*}{\textbf{Model}}}  & 
                \multicolumn{2}{c}{\textbf{IEMOCAP}}
                \\ \cmidrule[0.05em](rl){2-3}
                \multicolumn{1}{c}{} & 
                \multicolumn{1}{c}{\begin{tabular}[c]{@{}c@{}}Speaker Indep.\\(random) \end{tabular}} & \multicolumn{1}{c}{\begin{tabular}[c]{@{}c@{}}Speaker Overlap\\(random) \end{tabular}}
                \\ \midrule[0.05em]
                BLSTM~\cite{RW9}                                    & 58.81 & — \\
                Fully Convolutional Network~\cite{RW3}              & 63.91 & — \\
                CNN Transformer~\cite{RW8}                          & 64.79 & \textbf{70.05} \\
                CRNN with Transfer Learning \& Augm.~\cite{SOTA5}   & \textbf{65.02} & — \\
                \multicolumn{1}{l}{Ours (Centralized)}              & 62.58 & 69.04 \\
                \midrule[0.05em]
                \multicolumn{1}{l}{Ours (Federated)}                & 60.52 & 64.78 \\ 
                \bottomrule
            \end{tabular}%
        }
    \end{center}
\end{table}

\begin{figure}[b]
    \centerline{\includegraphics[width=0.78\columnwidth]{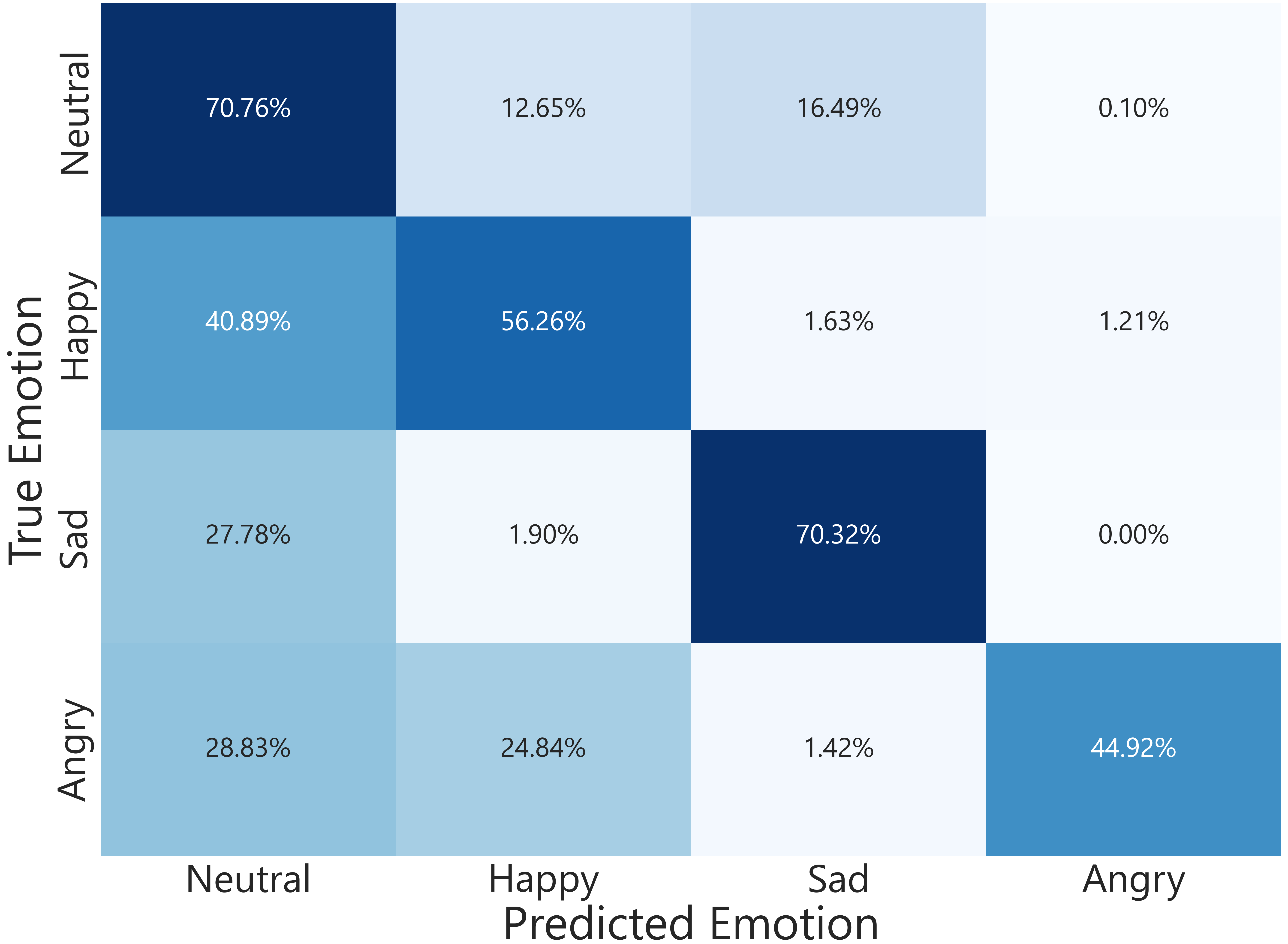}}
    \caption{Confusion Matrix of Speaker Independent Federated SER\label{fig:cfm}}
\end{figure}

\textit{\textbf{Centralized vs. Federated SER}}: Since, to the best of our knowledge, no federated learning-based SER exists, we compare our model with state-of-the-art centralized SER approaches to establish an upper bound of performance of our model and to determine the feasibility of performing SER in a federated setting. To this end, we use attention-based state-of-the-art architectures, such as recurrent neural networks (BLSTM and CRNN) and CNN's, which often rely on transfer learning and augmentations to improve their performance. However, these approaches require long training procedures and large-scale models. Thus, they are unfit for federated settings, where devices operate under low energy and computation constrains. We perform experiments in fully-supervised settings, both in a centralized and federated fashion, to construct a high-quality \textit{supervised} baseline. In federated settings, the FL parameters were set to $R$=100, $q$=80\%, $E$=1, $\sigma$=25\%,$L$=100\%, and $K$=10. The resulting unweighted accuracy on the test set for centralized and federated settings are presented in Table~\ref{tab:baseline}.\par

In Table~\ref{tab:baseline}, we observe that our model, despite its simplicity, is within 2.5\% from recent state-of-the art approaches, which rely on additional techniques (e.g., transfer learning, data augmentations) to improve their performance. We note that in Speaker Overlap experiments, the accuracy in both centralized and supervised federated settings is increased substantially, verifying that generalization is still an open research question in SER. Moving from centralized to federated settings, we notice that our model is able to retain high level of accuracy, with an accuracy gap between the centralized and federated models of approximately 4\% and 2\% for Speaker Independent and Speaker Overlap experiments, respectively.This indicates that utilizing FL to develop highly accurate privacy-preserving SER systems is feasible. Furthermore, to clearly demonstrate the performance of federated SER models, a confusion matrix for speaker independent experiments is presented in Figure~\ref{fig:cfm}. From the 4 emotions, we note that the \textit{angry} emotion has the largest misclassification rate. Considering that audio samples expressing an angry emotion are on average over 4.5 seconds long and since our model utilizes audio segments of two seconds, there might be difficulties to detect the particular emotional state.\par

\textit{\textbf{Effect of Low On-Device Label Data on Federated SER}}: We also evaluated performance of our SER model when on-device labeled data are scarce. For this, we employed federated self-training to determine the obtained improvements versus a fully-supervised federated approach, where we only utilized 10\% of the available data as labeled samples ($L$=10\%). To illustrate the performance gain of our federated self-training approach over the supervised FL regime, we performed fully-supervised experiments with identical labeled subsets, where, the unlabeled instances remained unexploited.\par

\begin{table}[t]
    \caption{{UA\% for Federated SER with 10\% of data containing labels}\label{tab:results}}
    \begin{center}
        \resizebox{\columnwidth}{!}{%
            \begin{tabular}{lcccc}
                \toprule
                \multicolumn{1}{c}{\multirow{4}{*}{\textbf{\begin{tabular}[c]{@{}c@{}}Federated\\Experiment\end{tabular}}}} &
                \multicolumn{1}{l}{\multirow{4}{*}{\textbf{Devices}}} &
                \multicolumn{3}{c}{\textbf{IEMOCAP}}\\
                \multicolumn{1}{c}{} & & 
                    \multicolumn{2}{c}{\textit{\textbf{Supervised}}} & 
                    \textit{\textbf{Semi-Supervised}}
                \\ \cmidrule[0.05em](lr){3-4} \cmidrule[0.05em](lr){5-5}
                \multicolumn{1}{c}{}            &   & $L$=100\% & $L$=10\% & $L$=10\% \\ 
                \midrule[0.05em]
                Speaker Indep. (random)         & 10 & 60.52 & 46.72 & \textbf{53.94} \\
                Speaker Indep. (non-i.i.d.)     &  8 & 59.55 & 46.55 & \textbf{56.83} \\
                Speaker Overlap (random)        & 10 & 64.78 & 51.58 & \textbf{59.04} \\ 
                Speaker Overlap (non-i.i.d.)    &  8 & 62.79 & 49.97 & \textbf{59.72} \\ 
                \bottomrule
            \end{tabular}%
        }
    \end{center}
\end{table}

In Table~\ref{tab:results}, comparing the two approaches for $L$=10\%, we note that our method utilize on-device unlabeled data to improve the model's performance significantly across all four experiments, with an average increase on accuracy by 8.67\%. Additionally, for the non-i.i.d. experiments, we notice that both supervised and semi-supervised models retain their accuracy, with the latter to improve the recognition rate by 1.75\% on average.
These results suggest that SER is feasible in federated settings, even under highly non-i.i.d. speakers distributions. If we recollect that in FL, such non-i.i.d. distributions pose significant challenges during training~\cite{FL_Challenges}, being unrestrained by those challenges is yet another reason to learn federated SER models and build privacy-preserving systems.\par

\section{Conclusion\label{sec:conclusion}}
We proposed a privacy-preserving SER model by utilizing Federated Learning. To eliminate the assumption of abundant labeled data availability on devices, we utilize a data-efficient federated self-training method to learn SER models with few on-device labeled samples. From our evaluation, we demonstrate that our models' accuracy is consistently superior to fully supervised federated settings under the same labeled data availability. In addition, contrary to other audio recognition tasks, we show that highly non-i.i.d. speaker distributions have minor effect in federated SER models’performance. \par

\section*{Acknowledgment}
The work presented in this paper is performed in the context of the DAIS project supported by the ECSEL Joint Undertaking (JU). JU receives support from the European Union's Horizon 2020 research and innovation program and Sweden, Netherlands, Germany, Spain, Denmark, Norway, Portugal, Belgium, Slovenia, Czech Republic, Turkey.

\bibliographystyle{IEEEtran}
\bibliography{main.bbl}

\end{document}